\documentclass[letterpaper, 10 pt, conference]{ieeeconf}  %
\IEEEoverridecommandlockouts       
\usepackage{amsmath}
\usepackage{amssymb}
\usepackage{booktabs}
\usepackage{multicol}
\usepackage{multirow}
\usepackage{graphicx}
\usepackage{color}
\usepackage{balance}

\title{MATrack: Efficient Multiscale Adaptive Tracker for Real-Time \\ Nighttime UAV Operations}

\author{
Xuzhao Li$^{*}$\thanks{*Equal contribution.} \ \
Xuchen Li$^{*}$ \ \
Shiyu Hu$^{\dagger}$\thanks{$\dagger$Corresponding authors.}
\\
Nanyang Technological University\\
xuzhaoli2001@gmail.com, xuchenli1030@gmail.com, shiyu.hu@ntu.edu.sg
}

\begin{document}

\maketitle
\thispagestyle{empty}
\pagestyle{empty}

\begin{abstract}
Nighttime UAV tracking faces significant challenges in real-world robotics operations. Low-light conditions not only limit visual perception capabilities, but cluttered backgrounds and frequent viewpoint changes also cause existing trackers to drift or fail during deployment. To address these difficulties, researchers have proposed solutions based on low-light enhancement and domain adaptation. However, these methods still have notable shortcomings in actual UAV systems: low-light enhancement often introduces visual artifacts, domain adaptation methods are computationally expensive and existing lightweight designs struggle to fully leverage dynamic object information. Based on an in-depth analysis of these key issues, we propose MATrack—a multiscale adaptive system designed specifically for nighttime UAV tracking. MATrack tackles the main technical challenges of nighttime tracking through the collaborative work of three core modules:  Multiscale Hierarchy Blende (MHB) enhances feature consistency between static and dynamic templates. Adaptive Key Token Gate accurately identifies object information within complex backgrounds. Nighttime Template Calibrator (NTC) ensures stable tracking performance over long sequences. Extensive experiments show that MATrack achieves a significant performance improvement. On the UAVDark135 benchmark, its precision, normalized precision and AUC surpass state-of-the-art (SOTA) methods by 5.9\%, 5.4\% and 4.2\% respectively, while maintaining a real-time processing speed of 81 FPS. Further tests on a real-world UAV platform validate the system's reliability, demonstrating that MATrack can provide stable and effective nighttime UAV tracking support for critical robotics applications such as nighttime search and rescue and border patrol.
\end{abstract}

\section{Introduction}

As a core task of modern robotic vision systems, unmanned aerial vehicle (UAV) object tracking plays an irreplaceable role in critical applications such as border patrol \cite{lei2023multi}, nighttime search and rescue \cite{al2019appearance} and aerial reconnaissance \cite{tian2011video}. This ability to automatically follow moving objects from an aerial platform provides vital technical support for real-world applications. However, as UAV operations expand into nighttime environments, traditional tracking technologies are facing unprecedented challenges. This operational shift is crucial for missions that require continuous surveillance capabilities. In recent years, single object tracking technology \cite{cao2023towards} has made significant progress, driven by advancements in deep learning \cite{he2016deep} and the Transformer architecture \cite{shen2024overlapped, vaswani2017attention}. Algorithms like MixFormer \cite{cui2022mixformer}, ODTrack \cite{zheng2024odtrack} and VideoTrack \cite{xie2023videotrack} have demonstrated excellent performance in daytime conditions by modeling global context and learning sequence-level representations. However, when these advanced techniques are applied to nighttime UAV tracking, their limitations are quickly exposed. The low-light conditions of nighttime environments cause a sharp drop in image signal-to-noise ratio, while frequent viewpoint changes and motion blur further degrade feature quality. Meanwhile, complex background clutter increases the risk of mistracking. The UAV's inherent motion, combined with these visual challenges, real-time requirements and resource constraints, creates a complex technical problem for stable nighttime object tracking.

\begin{figure}[t]
    \centering
    \includegraphics[width=0.48\textwidth]{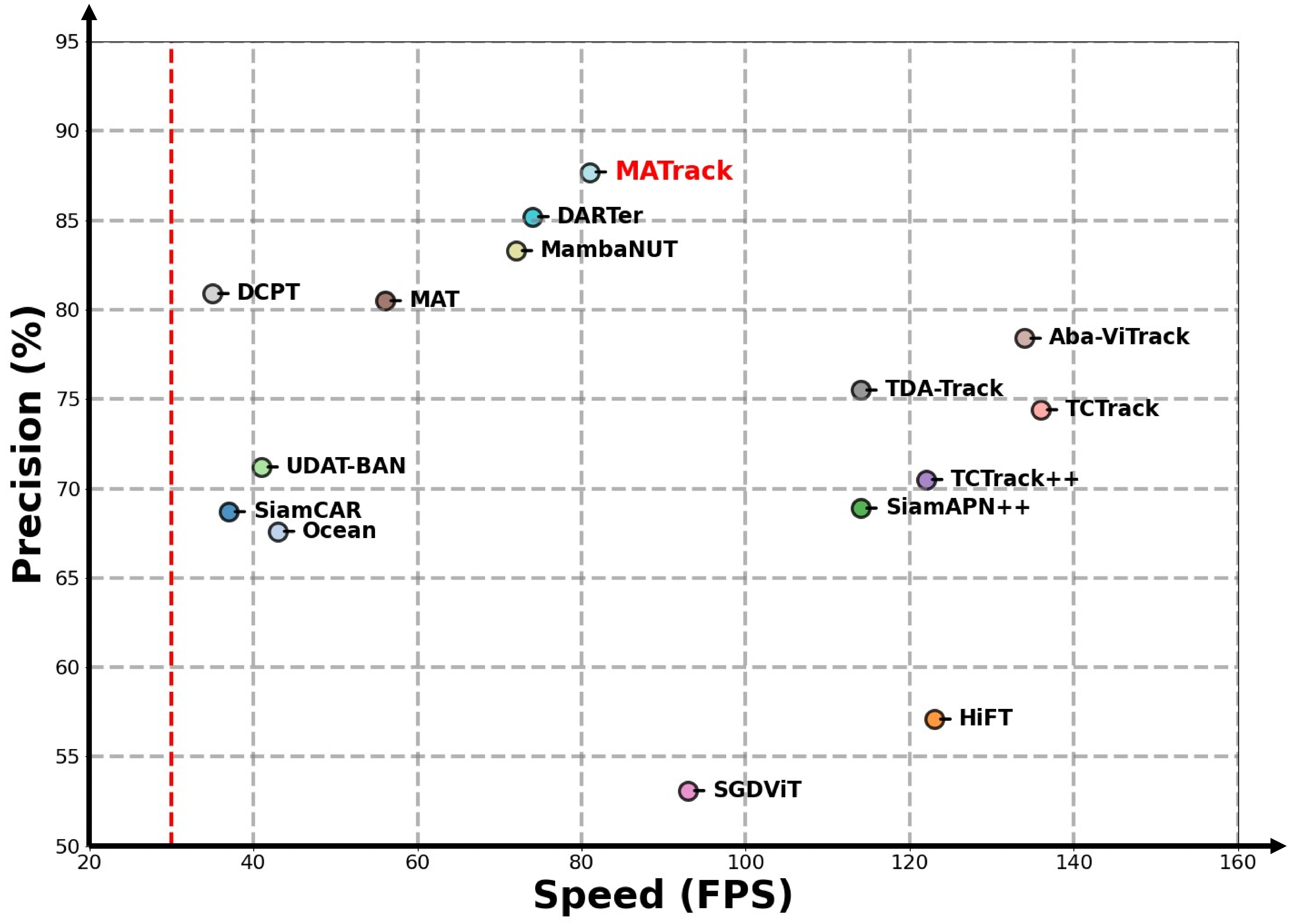}
    \caption{This scatter plot illustrates the balance between tracking speed (FPS) and precision (\%) for various methods on NAT2024-1 \cite{fu2024prompt} benchmark. The red dashed line marks the 30 FPS threshold for real-time performance. Our proposed MATrack (highlighted in red) achieves the highest precision (87.7\%) while maintaining a speed (81 FPS) well above the real-time requirement, demonstrating a superior efficiency-performance trade-off compared to other state-of-the-art (SOTA) methods.}
    \label{fig:bubble}
\end{figure}

To tackle this challenge, researchers explore three main directions. The first category of methods uses low-light enhancement techniques (e.g., HighlightNet \cite{fu2022highlightnet}, Darklighter \cite{ye2021darklighter}) to improve image quality. However, this pre-processing often introduces visual artifacts that, in the highly dynamic flight environment of an UAV, can interfere with tracking accuracy. For instance, boosting low-light signals can inadvertently amplify sensor noise or generate false edges, which confuse the tracker. The second category employs domain adaptation techniques (e.g., UDAT \cite{ye2022unsupervised}, SAM-DA \cite{fu2024sam}) to narrow the distribution gap between daytime and nighttime features. While these methods can improve tracking performance to some extent, their high training costs and computational demands are difficult to meet on resource-constrained UAV platforms. The offline nature of these methods also limits their adaptability to unforeseen real-world scenarios. The third category focuses on lightweight and efficient network designs (e.g., DCPT \cite{zhu2024dcpt}, MambaNUT \cite{wu2024mambanut}, DARTer \cite{li2025darter}) to improve computational efficiency, but they still fall short in adapting to dynamic environments, resisting noise and maintaining long-term stability. A common limitation of all these methods is that they primarily seek algorithmic solutions while overlooking the fundamental nature of nighttime UAV tracking as a system-level problem. The core issue is not just about a single-component solution but about creating a robust, end-to-end framework that addresses the holistic set of challenges.

Recognizing this, we propose MATrack—a multiscale adaptive tracking framework designed from a system perspective. MATrack integrates three synergistic core modules: the Multiscale Hierarchical Blende (MHB) enhances feature consistency and robustness by unifying static and dynamic template information; the Adaptive Key Token Gate (AKTG) dynamically identifies and strengthens object-related visual cues in complex nighttime environments; and the Nighttime Template Calibrator (NTC) ensures the stability of the tracking system over long sequences through an intelligent update mechanism. This collaborative design allows MATrack to generate highly discriminative object representations even under severe light degradation, while meeting the real-time and resource constraints of UAV platforms.

Extensive experiments fully validate MATrack's superiority. In a comprehensive evaluation across five nighttime tracking benchmarks, MATrack not only achieves state-of-the-art performance on all metrics—surpassing the best existing methods by 6.0\% in precision, 5.4\% in normalized precision and 4.2\% in AUC on UAVDark135 \cite{li2022all} benchmark—but also strikes an ideal balance between efficiency and performance with a speed of 81 FPS. More importantly, deployment tests on a real-world UAV platform have confirmed MATrack's practical utility, demonstrating that it is not merely an algorithmic advancement but a reliable engineering solution capable of operating effectively in real systems. This real-world validation confirms its robustness beyond theoretical benchmarks, proving its efficiency for practical deployment in missions like search and rescue or surveillance.

In summary, our contributions are as follows:
\begin{itemize}
    \item We propose the Multiscale Hierarchy Blender (MHB) which hierarchically fuses static and dynamic templates with the search region to enhance multiscale consistency and robustness.
    \item We introduce the Adaptive Key Token Gate (AKTG) to dynamically balances local and global feature cues, suppresses background noise and emphasizes object-related tokens.
    \item We design Nighttime Template Calibrator (NTC) module which adaptively updates dynamic templates through an offset-aware mechanism, ensuring reliable long-term tracking under challenging conditions.
    \item We achieved new state-of-the-art (SOTA) results on five benchmarks while maintaining real-time performance. Furthermore, we validated the system's practicality through real-world UAV deployment, demonstrating a complete chain from algorithmic innovation to practical application.
\end{itemize}

\section{Related Works}
\subsection{Single Object Tracking}
The purpose of single object tracking is to track a object in challenging scenarios such as those with similar object interference, occlusion and complex backgrounds. With the development of deep learning, MixFormer \cite{cui2022mixformer}, as a concise end-to-end model based on Transformer, relies on a backbone network that mixes the template and search images together with a regression head to directly output tracking results. ARTrack \cite{wei2023autoregressive,bai2024artrackv2} transforms tracking into a coordinate sequence interpretation task. OSTrack \cite{ye2022joint} adopts the ViT network architecture to build an efficient visual tracking framework. VideoTrack \cite{xie2023videotrack} integrates context information. Similarly, ODTrack \cite{zheng2024odtrack} proposes a token sequence propagation method to associate various types of context information. OTETrack \cite{shen2024overlapped} adds an additional template and continuously updates the additional template to provide more information. To reduce the complexity of the tracker, MCITrack \cite{kang2025exploring} uses the mamba and leverages its linear complexity optimization and long-sequence processing capabilities to build a new tracking framework. The EVPTrack \cite{shi2024explicit} uses a spatio-temporal encoder to propagate information between consecutive frames through tokens and combines a prompt generator to generate multiscale and spatio-temporal explicit visual prompts. LoReTrack \cite{dong2024loretrack} improves the tracking performance by enabling the low-resolution tracker to inherit the feature interaction of the high-resolution model from a global perspective.

\begin{figure*}[!t]
    \centering
    \vspace{2mm}
    \includegraphics[width=0.98\textwidth]{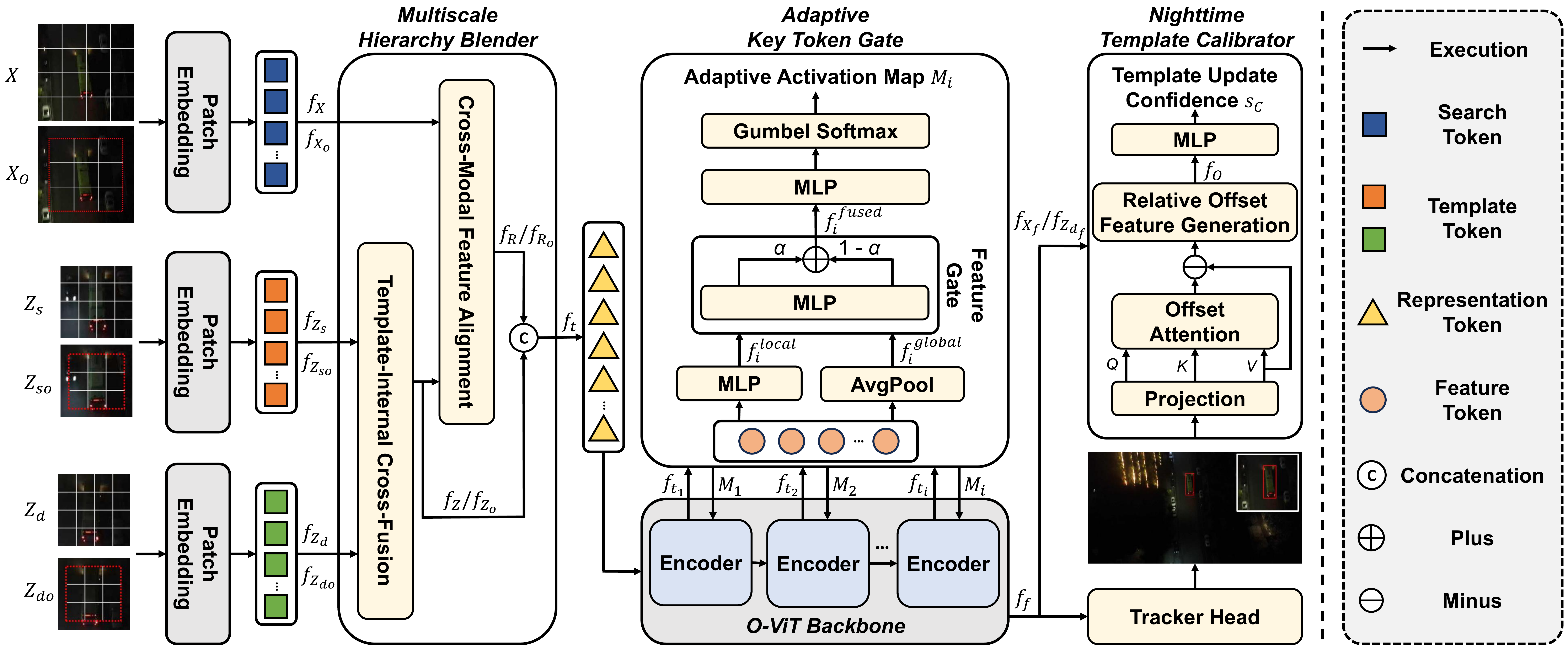}
    \caption{Overview architecture of MATrack. The nighttime dynamic features of the static, dynamic templates and research region are fused by Multiscale Hierarchy Blender (MHB) module. As the O-ViT backbone performs feature extraction and interaction, the Adaptive Key Token Gate (AKTG) suppresses background noise tokens while emphasizing object-related information. We also designed the Nighttime Template Calibrator (NTC) to adaptively update the dynamic template and ensure reliable long-term tracking under challenging conditions.}
    \label{fig:overview}
\end{figure*}

\subsection{Nighttime UAV Tracking}
Due to factors such as lower illumination conditions, nighttime UAV tracking is a much more challenging task. For light enhancement, Highlightnet \cite{fu2022highlightnet} uses a pixel-level range mask in its adaptive low-light enhancer to focus on targets, Darklighter \cite{ye2021darklighter} improves low-light image quality by estimating light and noise maps, Ye et al. \cite{ye2022tracker} train the SCT enhancer via task-inspired perceptual loss for denoising and light adjustment, and ADTrack \cite{li2021adtrack} combines a low-light enhancer with a correlation filter-based framework. Regarding domain adaptation, Fu et al. \cite{fu2024prompt} align day-night spatio-temporal contexts. SAM-DA \cite{fu2024sam} proposes a training framework. UDAT \cite{ye2022unsupervised} proposes the first unsupervised domain adaptation nighttime aerial tracking framework. DCPT \cite{zhu2024dcpt} learns visual prompts iteratively, and MambaNUT \cite{wu2024mambanut} leverages Vision Mamba's \cite{liu2024vmamba} linear complexity. DARTer \cite{li2025darter} is an end-to-end framework for nighttime UAV tracking that improves accuracy and efficiency by adaptively fusing multi-perspective features and activating Vision Transformer layers based on the scene's dynamics. However, these methods rely on extensive training, incur high costs, increase optimization complexity, and fail to fully exploit dynamic information from extreme viewpoint changes.  Compared to existing methods, MATrack is a robust, end-to-end system solution that uses three unique core modules. It effectively suppresses background noise and visual artifacts common in nighttime environments, achieving both high accuracy and real-time efficiency, as validated by real-world UAV deployment.

\section{Methodology}
\subsection{Overview}
We propose a nighttime UAV tracking framework, named MATrack. Its architecture is shown in Fig. \ref{fig:overview}. The process begins by taking search image, along with both static and dynamic templates, and slicing them into O-patches \cite{shen2024overlapped}. We use a Multiscale Hierarchy Blender (MHB) to align features of different scales from the static and dynamic templates and the search image. Subsequently, all these features are fed into the O-ViT \cite{shen2024overlapped}. Furthermore, we use an Adaptive Key Token Gate (AKTG) to suppress background noise tokens while enhancing object-related tokens, thus improving tracking performance. Simultaneously, we use a Nighttime Template Calibrator (NTC) to enable efficient and accurate template updates. The following sections will provide more details on these components.

\subsection{Multiscale Hierarchy Blender}
The input images of MATrack include the search images $X\in \mathbb{R}^{3 \times H_x \times W_x}$, the static template images $Z_s \in \mathbb{R}^{3 \times H_{z_s} \times W_{z_s}}$ and the dynamic template images $Z_d \in \mathbb{R}^{3 \times H_{z_d} \times W_{z_d}}$. We adopt the current ViT-based tracking paradigm \cite{ye2022joint}, partitioning images into patches and then converting them into token sequences. After processing, we obtain the  initial search features $f_X$, the static template features $f_{Z_s}$ and the dynamic template features $f_{Z_d}$. Meanwhile, these images are sliced into O-patches \cite{shen2024overlapped}, including $f_{X_o}$, $f_{Z_{so}}$ and $f_{Z_{do}}$, which enhances the correlation between image patches across different scales. 

To accurately capture object features at different scales, effectively filter isolated background noise and highlight consistent features between the object and templates, we use the Multiscale Hierarchy Blender (MHB) module to perform hierarchical feature fusion on the static template, dynamic template and search image. 
 
Specifically, we perform Template-Internal Cross-Fusion on the features of the initial static and dynamic templates (\( f_{Z_s} \) and \( f_{Z_d} \)), as well as their overlapped features (\( f_{Z_{s_o}} \) and \( f_{Z_{d_o}} \)). This process yields the primary blended features, \( f_Z \) and \( f_{Z_o} \). The calculation for the initial static and dynamic templates is as follows:
\begin{equation}
\label{equ:fzs}
\begin{aligned}
f_{Z_s'} &= \Phi_{CA}(f_{Z_s}, f_{Z_d}), \quad &f_{Z_{so}'} &= \Phi_{CA}(f_{Z_{so}}, f_{Z_{do}}), \\
f_{Z_d'} &= \Phi_{CA}(f_{Z_d}, f_{Z_s}), \quad &f_{Z_{do}'} &= \Phi_{CA}(f_{Z_{do}}, f_{Z_{so}}), \\
f_Z &= \text{Concat}(f_{Z_s'}, f_{Z_d'}), \quad &f_{Z_o} &= \text{Concat}(f_{Z_{so}'}, f_{Z_{do}'}),
\end{aligned}
\end{equation}
where $\Phi_{CA}$ represents the cross-attention operation and \text{Concat} represents the concatenation operation. In this operation, the first element functions as Q and the second element is used to acquire K and V \cite{vaswani2017attention}. 

Subsequently, we adopt Cross-Modal Feature Alignment to achieve multiscale feature alignment between the search feature $f_X$ with $f_{X_o}$ and the templates features $f_Z$ with $f_{Z_o}$, generating two cross-modal interactive feature representations $f_{R}$ and $f_{R_o}$. This provides a more comprehensive feature foundation for subsequent matching.
\begin{equation}
\label{equ:fr}
\begin{aligned}
f_{R} = \Phi_{CA}(f_X, f_Z), \quad 
f_{R_o} = \Phi_{CA}(f_{X_o}, f_{Z_o}).
\end{aligned}
\end{equation}

Finally, we perform global feature integration. All fused multiscale features are concatenated to form a feature representation $f_t$ that contains the global information of both the search frame and the templates:
\begin{equation}
\label{equ:ft}
\begin{aligned}
f_{t} = \text{Concat}(f_{R}, f_{R_o}, f_Z, f_{Z_o}).
\end{aligned}
\end{equation}

\subsection{Adaptive Key Token Gate}

We propose the Adaptive Key Token Gate (AKTG) module. This module calculates the Adaptive Activation Map based on the fused features from the previous O-ViT block \cite{shen2024overlapped} combined with the feature gate, dynamically adjusts the attention to local and global information, and then suppresses background noise tokens and emphasizes object-related tokens through attention correction.

First, the AKTG module performs fine-grained feature splitting and sub-feature extraction on each output feature $f_{t_i}$ from the $i$-th O-ViT. These features are split by the number of attention heads $h$, into sub-features $f_i$, with each attention head processing a sub-feature independently.

Then, a dual-path feature extraction is performed on each sub-feature $f_i$. Specifically, we perform local nighttime feature extraction to capture single-token details, denoted as $f_i^{local}$:
\begin{equation}
\label{equ:filocal}
\begin{aligned}
f_i^{local} = \text{MLP}(f_i).
\end{aligned}
\end{equation}

Simultaneously, we extract global nighttime features to capture overall contextual relationships, denoted as $f_i^{global}$:
\begin{equation}
\label{equ:figlobal}
\begin{aligned}
f_i^{global} = \text{AvgPool}(\text{MLP}(f_i)).
\end{aligned}
\end{equation}

To address the unreliability of local details and the robustness of global information in nighttime UAV tracking, we propose the feature gate mechanism to adaptively weigh and fuse local and global features, which handles complex and changing nighttime environments.

Specifically, we input $f_i^{local}$ and $f_i^{global}$ into the feature gate to obtain the activation weights $\alpha$:
\begin{equation}
\label{equ:alpha}
\begin{aligned}
\alpha = \text{MLP}(\text{Concat}(f_i^{local}, f_i^{global})).
\end{aligned}
\end{equation}

Using the predicted activation weights $\alpha$, we perform a weighted sum of the local and global features to obtain the final fused feature $f_{i}^{fused}$:
\begin{equation}
\label{equ:ffused}
\begin{aligned}
f_{i}^{fused} = \alpha \odot f_{i}^{local} + (1 - \alpha) \odot f_{i}^{global},
\end{aligned}
\end{equation}
where $\odot$ denotes the element-wise multiplication. 

We then employ Gumbel-Softmax \cite{xu2022groupvit} to generate the Adaptive Activation Map $M_i \in \{0 \sim 1\}^N$:
\begin{equation}
\label{equ:mi}
\begin{aligned}
M_i = \text{GumbelSoftmax}(\text{MLP}(f_{i}^{fused})).
\end{aligned}
\end{equation}

In complex nighttime scenarios, we dynamically adjust the focus on local and global information using the Adaptive Activation Map $M_i$. 

To continuously suppress background noise tokens and emphasize object-related tokens, i.e. key tokens, we apply attention correction to the attention map $A_{map_i}$ within the O-ViT block:
\begin{equation}
\label{equ:ac}
\begin{aligned}
\text{AC}_{map_i} = (A_{map_i} \cdot M_i + A_{map_i}) \cdot V_i,
\end{aligned}
\end{equation}
where $V_i$ is the value matrix from $i$-th O-ViT block.

\subsection{Nighttime Template Calibrator}
In complex nighttime environments, previous trackers often rely on fixed time intervals or simple thresholds for dynamic template updates \cite{li2025darter}, which can easily lead to low-quality or even invalid dynamic templates and in turn reduce tracking accuracy and efficiency.
To address this challenge, we propose the Nighttime Template Calibrator (NTC) module, which performs dynamic template calibration through an offset-aware mechanism.

From the output of the final O-ViT block $f_f$, we partition it by index into $f_{X_f}$, $f_{Z_f}$ and $f_{Z_{d_f}}$. We then use Offset-Attention to compute the relative offset between the dynamic template and the search frame.

We map the features to $Q_n$, $K_n$ and $V_n$ matrices:
\begin{equation}
\label{equ:qkv}
\begin{aligned}
Q_n &= \Phi_p(f_{Z_{d_f}}),\\
K_n &= \Phi_p(f_{X_f}), \\
V_n &= \Phi_p(f_{X_f}),
\end{aligned}
\end{equation}
where $\Phi_p$ represents the projection operation.

After that, we perform offset attention calculation, which is computed as follows:
\begin{equation}
\label{equ:att}
\begin{aligned}
Attention(Q_n, K_n, V_n) = \text{Softmax}\left( \frac{Q_n \cdot K_n^T}{\sqrt{d_k}} \right) \cdot V_n ,
\end{aligned}
\end{equation}
where $d_k$ is the dimension of the $Q_n$ and $K_n$ vectors.

We generate the relative offset feature $f_O$ and gain the template update confidence $s_c \in (0,1)$:
\begin{equation}
\label{equ:fo}
\begin{aligned}
f_O = \text{ReLU}\left( \text{InsNorm}\left(\Phi_l(V_n - \text{Attention}(Q_n, K_n, V_n)) \right) \right),
\end{aligned}
\end{equation}
where $\Phi_l$ represents a linear transformation layer and $\text{InsNorm}$ is a instance normalization operation.
\begin{equation}
\label{equ:sc}
\begin{aligned}
s_c = \text{MLP}(f_O).
\end{aligned}
\end{equation}

Let $\theta$ be the confidence score threshold. If $s_c \in \theta$, we update the dynamic template.

\subsection{Prediction Head and Training Loss}
Following the architecture of models such as MixFormer \cite{cui2022mixformer} and DARTer \cite{li2025darter}, we utilize a prediction head comprising four stacked Conv-BN-ReLU layers. This head first transforms the output tokens of the search image into a 2D spatial feature map. It then processes these features to output three distinct results for each potential object. The final bounding box is located at the position with the peak classification score.

For the training, MATrack's loss function, $L_{\text{total}}$, is a weighted combination of the softmax cross-entropy loss ($L_{ce}$) \cite{wei2023autoregressive} and the SloU loss ($L_{SloU}$) \cite{gevorgyan2022siou}, given by the formula $L_{\text{total}}=\lambda_{1}L_{ce}+\lambda_{2}L_{SloU}$. Both weights ($\lambda_1$ and $\lambda_2$) were set to 2 in our experiments.

\begin{table*}[t]
\centering
\vspace{2mm}
\caption{State-of-the-art comparison on the NAT2024-1 \cite{fu2024prompt}, NAT2021 \cite{ye2022unsupervised} and UAVDark135 \cite{li2022all} benchmarks. The top three results are highlighted in \textcolor{red}{\textbf{red}}, \textcolor{blue}{\textbf{blue}} and \textcolor{green}{\textbf{green}}, respectively. Note that the percent symbol (\%) is excluded for precision score (P), normalized precision (P\(_\text{Norm}\)) and area under the curve (AUC).}
\label{tab:overall_performance}
\resizebox{6.9in}{!}{
\begin{tabular}{c|c|ccc|ccc|ccc} 
\toprule
\multirow{2}{*}{Tracker} & \multirow{2}{*}{Source} & \multicolumn{3}{c|}{NAT2024-1} & \multicolumn{3}{c|}{NAT2021} & \multicolumn{3}{c}{UAVDark135}    \\
& & P & P\(_{\text{Norm}}\) & AUC & P & P\(_{\text{Norm}}\) & AUC & P & P\(_{\text{Norm}}\) & AUC  \\
\hline
SiamCAR \cite{guo2020siamcar} & CVPR 20 & 68.7	& 62.6	& 51.2	& 65.8	& 59.5	& 45.7	& 65.8	& 65.7	& 52.3  \\
Ocean \cite{zhang2020ocean} & ECCV 20 & 67.6	& 50.3	& 44.0	& 58.1	& 49.9	& 38.6	& 60.1	& 58.9	& 47.3  \\
HiFT \cite{cao2021hift} & ICCV 21 & 57.1 & 44.5 & 40.8 & 54.5 & 46.7 & 37.0 & 44.8 & 45.2 & 35.3  \\
SiamAPN++ \cite{cao2021siamapn++} & IROS 21 & 68.9 & 57.9 & 47.8 & 60.2 & 51.4  & 41.2 & 42.7 & 41.6 & 33.5  \\
UDAT-BAN \cite{ye2022unsupervised} & CVPR 22 & 71.2 & 64.9 & 51.1 & 68.9 & 58.8  & 47.2 & 61.1 & 61.7 & 48.4  \\
UDAT-CAR \cite{ye2022unsupervised} & CVPR 22 & 68.1 & 61.6 & 49.6 & 68.2 & 61.3 & 48.7 & 60.9 & 61.3 & 48.6  \\
TCTrack \cite{Cao2022TCTrackTC} & CVPR 22 & 74.4 & 51.2 & 47 & 60.8 & 51.9 & 40.8 & 49.8 & 50.0 & 37.7  \\
TCTrack++ \cite{cao2023towards} & TPAMI 23 & 70.5 & 50.8 & 46.6 & 61.1 & 52.8 & 41.7 & 47.4 & 47.4 & 37.8  \\
MAT \cite{zhao2023representation} & CVPR 23 & 80.5 & 76.3 & 61.9 & 64.8 & 58.8 & 47.7 & 57.2 & 57.6 & 47.1  \\
HiT-Base \cite{Kang2023ExploringLH} & ICCV 23 & 62.7 & 56.9 & 48.2 & 49.3 & 44.2 & 36.4 & 48.9 & 48.7 & 41.1  \\
Aba-ViTrack \cite{li2023adaptive} & ICCV 23 & 78.4 & 72.2 & 60.1 & 60.4 & 57.3 & 46.9 & 61.3 & 63.5 & 52.1  \\
SGDViT \cite{yao2023sgdvit} & ICRA 23 & 53.1 & 47.2 & 38.1 & 53.1 & 47.9 & 37.5 & 40.2 & 40.6 & 32.7  \\
TDA-Track \cite{fu2024prompt} & IROS 24 & 75.5 & 53.3 & 51.4 & 61.7 & 53.5 & 42.3 & 49.5 & 49.9 & 36.9  \\
AVTrack-DeiT \cite{lilearningicml} & ICML 24 & 75.3 & 68.2 & 56.7 & 61.5 & 55.6 & 45.5 & 58.6 & 59.2 & 47.6  \\
DCPT \cite{zhu2024dcpt} & ICRA 24 & 80.9  & 75.4 & 62.1 & 69.0 & 63.5 & \textcolor{green}{\textbf{52.6}} & 69.2 & \textcolor{green}{\textbf{69.8}} & 56.7  \\
MambaNUT \cite{wu2024mambanut} & IROS 25 & \textcolor{green}{\textbf{83.3}} & \textcolor{green}{\textbf{76.9}} & \textcolor{green}{\textbf{63.6}} & \textcolor{green}{\textbf{70.1}} & \textcolor{blue}{\textbf{64.6}} & 52.4 & \textcolor{green}{\textbf{70.0}} & 69.3 & \textcolor{green}{\textbf{57.1}}    \\
DARTer \cite{li2025darter} & ICMR 25 & \textcolor{blue}{\textbf{85.2}} & \textcolor{blue}{\textbf{80.1}} & \textcolor{blue}{\textbf{65.6}} & \textcolor{blue}{\textbf{70.2}} & \textcolor{green}{\textbf{63.7}} & \textcolor{blue}{\textbf{53.2}} & \textcolor{blue}{\textbf{71.6}} & \textcolor{blue}{\textbf{72.1}} & \textcolor{blue}{\textbf{58.2}} \\
\textbf{MATrack} & \textbf{Ours} & \textcolor{red}{\textbf{87.7}} & \textcolor{red}{\textbf{82.7}} & \textcolor{red}{\textbf{68.0}} & \textcolor{red}{\textbf{72.1}} & \textcolor{red}{\textbf{65.9}} & \textcolor{red}{\textbf{54.6}} & \textcolor{red}{\textbf{77.5}} & \textcolor{red}{\textbf{77.5}} & \textcolor{red}{\textbf{62.4}}
\\
\bottomrule
\end{tabular}}
\end{table*}

\section{Experiment}
\subsection{Implementation Details}
\textbf{Models.} We use Overlapped ViT \cite{shen2024overlapped} as the backbone. The head of MATrack consists of a stack of four
Conv-BN-Relu layers. The confidence score threshold is $\theta \in (0.3, 0.8)$. The image sizes of the search and template are $128 \times 128$ and $256 \times 256$, respectively. The initial and O patches of the search image are $16 \times 16$ and $15 \times 15$, and the initial and O patches of the template are $8 \times 8$ and $7 \times 7$, respectively. 

\textbf{Training.} For training, we used four common datasets: LaSOT \cite{fan2019lasot}, GOT10K \cite{huang2019got}, COCO \cite{lin2014microsoft}, and TrackingNet \cite{muller2018trackingnet}. Additionally, we incorporated three nighttime datasets—BDD100K\_Night \cite{yu2020bdd100k}, SHIFT\_night \cite{sun2022shift}, and ExDark \cite{loh2019getting}—to address low-light conditions. The model is trained for 150 epochs using the AdamW optimizer \cite{loshchilov2017decoupled}, with a batch size of 32. Each epoch involves 60,000 sampling pairs. The initial learning rate is set to 0.0001, and after 120 epochs, the learning rate decays at a rate of 10\%. The model is trained on a server with four A40 GPUs.

\textbf{Evaluation.} We evaluate MATrack on five mainstream benchmarks, including NAT2024-1 \cite{fu2024prompt}, NAT2021 \cite{ye2022unsupervised}, UAVDark135 \cite{li2022all}, NAT2021-L \cite{ye2022unsupervised} and DarkTrack2021 \cite{ye2022tracker}. We compare MATrack with the state-of-the-art (SOTA) trackers. All evaluation experiments are conducted on an RTX-4090 GPU.

\begin{table}[htbp!]
\centering
\caption{Comparison on the NAT2021-L \cite{ye2022unsupervised} benchmark. The top three results are highlighted in \textcolor{red}{\textbf{red}}, \textcolor{blue}{\textbf{blue}} and \textcolor{green}{\textbf{green}}, respectively.}
\label{tab:nat2021l_performance}
\resizebox{3.3in}{!}{
\begin{tabular}{c|c|cccc} 
\toprule
\multirow{2}{*}{Tracker} & \multirow{2}{*}{Source} & \multicolumn{3}{c}{NAT2021-L} \\
& & P & P\(_{\text{Norm}}\) & AUC\\
\hline
SiamRPN++ \cite{li2019siamrpn++} & CVPR 19 & 42.9 & 35.8 & 30.0 \\
Ocean \cite{zhang2020ocean} & ECCV 20 & 45.1 & 40.0 & 31.6 \\
HiFT \cite{cao2021hift} & ICCV 21 & 43.0 & 33.0 & 28.8 \\
SiamAPN \cite{fu2021siamese} & ICRA 21 & 37.7 & 27.7 & 24.2 \\
SiamAPN++ \cite{cao2021siamapn++} & IROS 21 & 40.0 & 32.7 & 28.0 \\
UDAT-BAN \cite{ye2022unsupervised} & CVPR 22 & 49.4 & 43.7 & 35.3 \\
UDAT-CAR \cite{ye2022unsupervised} & CVPR 22 & 50.4 & 44.7 & 37.8 \\
DCPT \cite{zhu2024dcpt} & ICRA 24  & \textcolor{green}{\textbf{58.6}} & \textcolor{green}{\textbf{54.6}} & \textcolor{green}{\textbf{47.4}} \\
DARTer \cite{li2025darter} & ICMR 25 & \textcolor{blue}{\textbf{64.9}} & \textcolor{blue}{\textbf{58.6}} & \textcolor{blue}{\textbf{50.9}} \\
\textbf{MATrack} & \textbf{Ours} & \textcolor{red}{\textbf{67.7}} & \textcolor{red}{\textbf{60.8}} & \textcolor{red}{\textbf{52.5}} \\
\bottomrule
\end{tabular}}
\end{table}

\subsection{Comparison Results}
\textbf{NAT2024-1.} NAT2024-1 \cite{fu2024prompt} focuses on simulating long-sequence tracking tasks in real-world low-illumination nighttime scenarios, addressing the insufficiency of existing benchmarks in covering long-term nighttime tracking scenarios. It comprises 40 long-term image sequences, with a total of more than 70,000 frames. As shown in Tab. \ref{tab:overall_performance}, MATrack achieves the best performance across all three metrics on NAT2024-1, it has a precision (P) score of 87.7\%, a normalized precision (P\(_{\text{Norm}}\)) of 82.7\% and an area under the curve (AUC) of 68.0\%. These results surpass the second-best tracker DARTer \cite{li2025darter} by 2.5\%, 2.6\% and 2.4\%, which confirms the robustness of MATrack on challenging UAV tracking sequences.

\textbf{NAT2021.} NAT2021 \cite{ye2022unsupervised} is specifically designed for tracking tasks in nighttime scenarios. It fills the gap of evaluation data in the field of nighttime UAV tracking and covers multiple dimensions including objects, environments and illumination. On NAT2021, MATrack achieves P, P\(_{\text{Norm}}\) and AUC scores of 72.1\%, 65.9\% and 54.6\%, respectively, outperforming all existing trackers. This highlights MATrack’s ability to maintain consistent accuracy even when objects undergo significant variations.

\textbf{NAT2021-L.} NAT2021-L \cite{fu2024prompt} is a long-term tracking benchmark that provides sufficient nighttime tracking videos for evaluating the performance of nighttime UAV tracking algorithms in long-term tracking scenarios, with each sequence containing more than 1,400 frames. On the NAT2021-L benchmark, MATrack records 67.7\% in P, 60.8\% in P\(_{\text{Norm}}\) and 52.5\% in AUC, ranking first across all metrics, showing that our tracker is much less affected by error accumulation in long nighttime scenarios.

\begin{table}[htbp!]
\centering
\caption{Comparison on the DarkTrack2021 \cite{ye2022tracker} benchmark. The top three results are highlighted in \textcolor{red}{\textbf{red}}, \textcolor{blue}{\textbf{blue}} and \textcolor{green}{\textbf{green}}, respectively.}
\label{tab:darktrack2021_performance}
\resizebox{3.3in}{!}{
\begin{tabular}{c|c|cccc}  
\toprule
\multirow{2}{*}{Tracker} & \multirow{2}{*}{Source} & \multicolumn{3}{c}{DarkTrack2021} \\
& & P & P\(_{\text{Norm}}\) & AUC\\
\hline
SiamRPN \cite{li2018high} & CVPR 18 & 50.9 & 48.5 & 38.7 \\
DIMP18 \cite{bhat2019learning} & ICCV 19 & 62.0 & 58.9 & 47.1 \\
PRDIMP50 \cite{danelljan2020probabilistic} & CVPR 20& 58.0 & 55.9 & 46.4 \\
SiamAPN++ \cite{cao2021siamapn++} & IROS 21 & 48.9 & 46.1 & 37.7 \\
HiFT \cite{cao2021hift} & ICCV 21  & 50.3 & 47.1 & 37.4 \\
SiamAPN++-SCT \cite{ye2022tracker} & RAL 22 & 53.7 & 51.1 & 40.8 \\
DIMP50-SCT \cite{ye2022tracker} & RAL 22 & \textcolor{blue}{\textbf{67.7}} & 63.3 & 52.1 \\
DCPT \cite{zhu2024dcpt} & ICRA 24 & \textcolor{blue}{\textbf{67.7}} & \textcolor{green}{\textbf{64.6}} & \textcolor{green}{\textbf{54.0}} \\
DARTer \cite{li2025darter} & ICMR 25 & \textcolor{green}{\textbf{67.6}} & \textcolor{blue}{\textbf{64.8}} & \textcolor{blue}{\textbf{54.5}} \\
\textbf{MATrack} & \textbf{Ours} & \textcolor{red}{\textbf{73.1}} & \textcolor{red}{\textbf{70.2}} & \textcolor{red}{\textbf{58.6}} \\
\bottomrule
\end{tabular}}
\end{table}

\textbf{UAVDark135.} UAVDark135 \cite{li2022all} is a benchmark specifically built for UAV nighttime tracking tasks. It defines a variety of common challenging attributes and consists of 135 sequences, with a total frame count of 125,466. On UAVDark135, MATrack demonstrates superior adaptability to low-light environments with a P of 77.5\%, a P\(_{\text{Norm}}\) of 77.5\% and an AUC of 62.4\%. Compared to DARTer, MATrack surpasses the SOTA tracker by 6.0\%, 5.4\% and 4.2\% in P, P\(_{\text{Norm}}\), and AUC, respectively, suggesting that its design generalizes well to extreme lighting conditions.

\textbf{DarkTrack2021.} DarkTrack2021 \cite{ye2022tracker} provides comprehensive evaluation support for nighttime UAV tracking algorithms. It contains a total of 110 challenging sequences and covers a rich variety of real-world nighttime UAV tracking scenarios. On DarkTrack2021, MATrack reaches the SOTA level in P, P\(_{\text{Norm}}\) and AUC, delivering a strong lead over all trackers. For instance, compared to DCPT and DARTer, MATrack improves by over 5\% in precision and nearly 4\% in AUC. These gains highlight MATrack’s resilience under both low illumination and cluttered backgrounds, where most trackers tend to fail due to noisy features.

\subsection{Efficiency analysis}
As demonstrated in Tab. \ref{tab:fps_performance4}, MATrack achieves a promising trade-off between speed and model size. Specifically, MATrack runs at 81 FPS, which is substantially faster than recent high-parameter trackers such as DCPT \cite{zhu2024dcpt}, while maintaining comparable parameter scale. Compared with lightweight trackers such as MambaNUT \cite{wu2024mambanut}, MATrack achieves significantly higher FPS than most heavy architectures and provides a stronger balance between efficiency and representational capacity.

\begin{table}[htbp!]
\centering
\caption{Comparison of our tracker and other state-of-the-art trackers in terms of average FPS and Parameters (M).}
\label{tab:fps_performance4}
\resizebox{3.3in}{!}{
\begin{tabular}{c|c|c|c} 
\toprule
\multirow{2}{*}{Tracker} & \multirow{2}{*}{Source} & \multirow{2}{*}{Average FPS} & \multirow{2}{*}{Parameters}  \\
& &  \\
\hline
SiamCAR \cite{guo2020siamcar} & CVPR 20 & 37 & 51.3 \\
Ocean \cite{zhang2020ocean} & ECCV 20 & 43	& 25.8 \\
HiFT \cite{cao2021hift} & ICCV 21 & 123 & 9.9 \\
SiamAPN++ \cite{cao2021siamapn++} & IROS 21 & 114 & 14.7 \\
UDAT-BAN \cite{ye2022unsupervised} & CVPR 22 & 41 & 54.1 \\
UDAT-CAR \cite{ye2022unsupervised} & CVPR 22 &36 & 54.6 \\
TCTrack \cite{Cao2022TCTrackTC} & CVPR 22 & 136 & 8.5 \\
TCTrack++ \cite{cao2023towards} & TPAMI 23 & 122 & 8.8 \\
MAT \cite{zhao2023representation} & CVPR 23 & 56 & 88.4 \\
HiT-Base \cite{Kang2023ExploringLH} & ICCV 23 & 156 & 42.1 \\
Aba-ViTrack \cite{li2023adaptive} & ICCV 23 & 134 & 7.9 \\
SGDViT \cite{yao2023sgdvit} & ICRA 23 & 93 & 23.3 \\
TDA-Track \cite{fu2024prompt} & IROS 24 & 114 & 9.2 \\
AVTrack-DeiT \cite{lilearningicml} & ICML 24 & 212 & 7.9 \\
DCPT \cite{zhu2024dcpt} & ICRA 24 & 35 & 92.9 \\
MambaNUT \cite{wu2024mambanut} & IROS 25 & 72 & 4.1   \\
DARTer \cite{li2025darter} & ICMR 25 & 74 & 80.9 \\
\textbf{MATrack} & \textbf{Ours} & 81 & 76.2 \\
\bottomrule
\end{tabular}}
\end{table}

\subsection{Ablation Study}
To verify the effectiveness of the proposed modules, we introduce them to the baseline tracker incrementally and provide experimental results on the NAT2024-1 \cite{fu2024prompt} dataset.

\begin{table}[htbp!]
\centering
\caption{Ablation studies on nighttime UAV tracking benchmarks NAT2024-1 \cite{fu2024prompt}.}
\resizebox{3.3in}{!}{
\begin{tabular}{c|ccc} 
\toprule
		Method & P & P\(_{\text{Norm}}\) & AUC  \\
		\hline
        {Base} & 84.1 & 79.6 & 65.1 \\
        {Base+MHB} & 85.8 (1.7$\uparrow$) & 80.8  (1.2$\uparrow$) & 66.2 (1.1$\uparrow$) \\
        {Base+MHB+AKTG} & 86.9 (2.8$\uparrow$) & 81.6 (2.0$\uparrow$) & 67.0 (1.9$\uparrow$) \\
        {Base+MHB+AKTG+NTC} & 87.7 (3.6$\uparrow$) & 82.7 (3.1$\uparrow$) & 68.0 (2.9$\uparrow$)  \\
		\bottomrule
\end{tabular}}
\label{tab:power_of_mm}
\end{table}

\textbf{Base+MHB.} The MHB module is the core of our tracker. It combines the stability of static templates with the dynamism of dynamic templates through multiscale feature fusion. Simultaneously, it aligns the features of the search frame with the templates at different scales, which boosts the robustness of the feature representation. As shown in Tab.\ref{tab:power_of_mm}, the incorporation of the MHB module resulted in a 1.1\% increase of AUC. 

\textbf{Base+MHB+AKTG.} We further add the AKTG module. As shown in Tab. \ref{tab:power_of_mm}, our MATrack achieves an AUC of 86.9\%, surpassing the baseline by 2.8\%. This demonstrates that the AKTG module provides robust, noise-resistant tracking by using its feature gate mechanism to adaptively focus on the object and suppress background noise.

\textbf{Base+MHB+AKTG+NTC} We further introduced the NTC module to achieve efficient and accurate dynamic template updates through its offset-aware capability.  As detailed in Tab. \ref{tab:power_of_mm}, the NTC module outperforms the baseline on the AUC, P and P\(_{\text{Norm}}\) metrics, showing the effectiveness of our proposed module.

\begin{figure}[htbp!]
    \centering
    \includegraphics[width=0.45\textwidth]{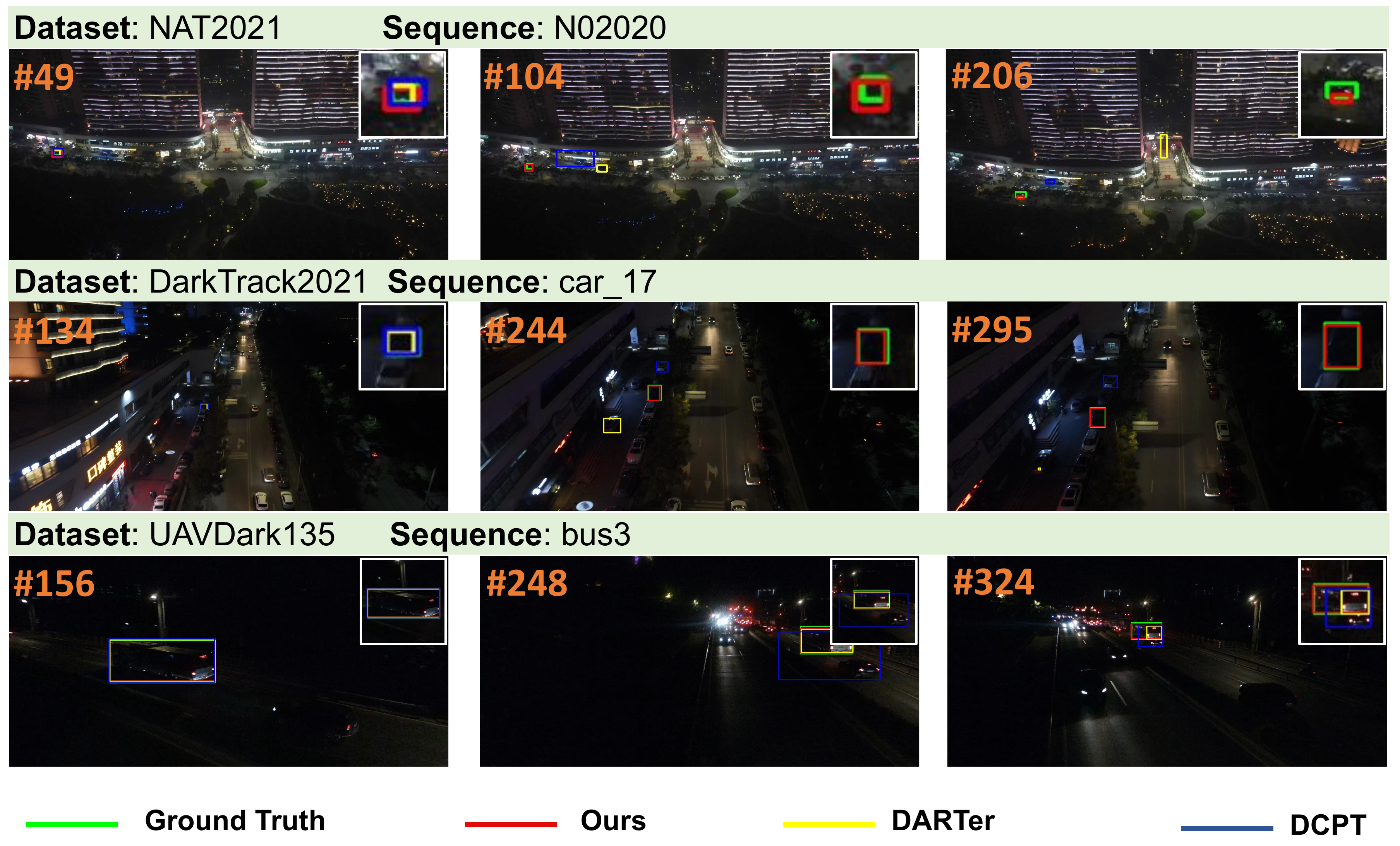}
    \caption{Qualitative comparison results of our tracker with other two latest trackers (i.e., DCPT \cite{zhu2024dcpt} and DRATer \cite{li2025darter}) in representative nighttime scenarios. Our method maintains its robustness even in complex environments. Better viewed in color with zoom-in.}
    \label{fig:vis}
\end{figure}

\subsection{Qualitative Analysis}
As shown in Fig.~\ref{fig:vis}, we visualize the tracking results of our model and the previous two SOTA models on three challenging sequences from NAT2021 \cite{ye2022unsupervised}, DarkTrack2021 \cite{ye2022tracker} and UAVDark135 \cite{li2022all}. In these sequences, the scenes contain distractors, and the state of the object undergoes significant changes. It is evident that our model exhibits greater robustness compared to others. This validates that our method contributes to addressing these challenges, further demonstrating the efficiency of our proposed method.

\begin{figure}[htbp!]
    \centering
    \includegraphics[width=0.45\textwidth]{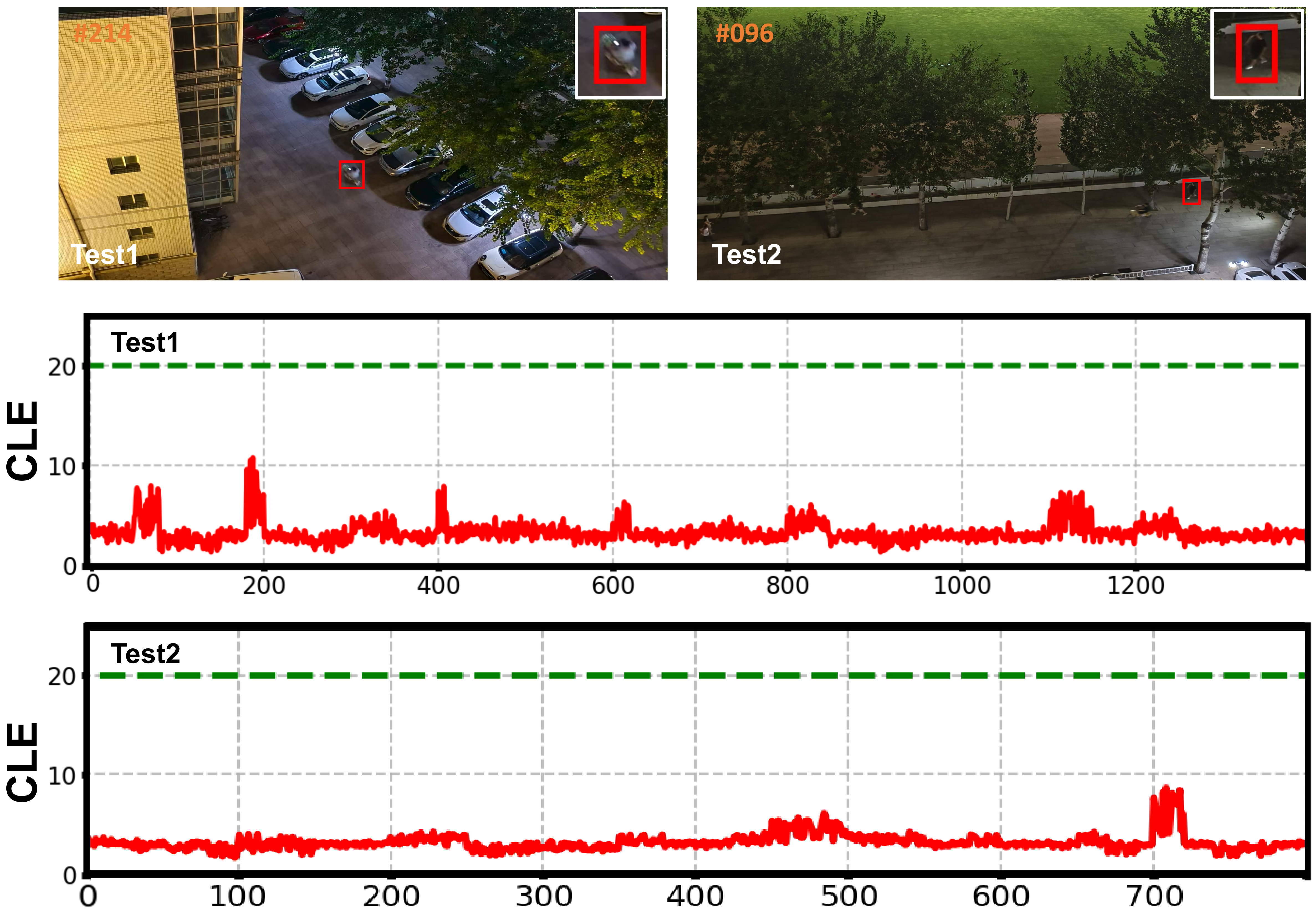}
    \caption{The reliability of our system is validated through real-world UAV platform tests in nighttime tracking scenarios. The frame-wise performance, represented by Center Location Error (CLE) plots, demonstrates that our tracker’s errors are consistently below the green dashed line (CLE = 20 pixels), which is the threshold for acceptable performance.}
    \label{fig:realworld}
\end{figure}

\section{Real-World Testing}
As shown in Fig. \ref{fig:realworld}, we conduct real-world tests to verify the performance of MATrack. We use the on-board camera on an UAV to capture nighttime images, and transmit these images to a computer in real time via Wi-Fi communication. The computer is equipped with an Nvidia 2080ti GPU, which can process the received images at a speed of over 30 FPS. The main challenges of these scenarios include viewpoint changes, partial occlusions and background noise. However, MATrack still achieve excellent performance, with the Center Location Error (CLE) of all test frames maintained below 20 pixels. Real-world tests show that MATrack is highly suitable for edge deployment on UAV platforms, delivering robust tracking performance in complex nighttime environments.

\section{Conclusion}
We introduce MATrack, a multiscale adaptive tracker that addresses the challenges of nighttime UAV tracking. By combining a Multiscale Hierarchy Blender for robust feature fusion, an Adaptive Key Token Gate for noise-resistant feature selection and a Nighttime Template Calibrator for dynamic template updates, MATrack sets a new state of the art. Extensive experiments show that MATrack consistently outperforms leading trackers in accuracy and robustness. Most importantly, it strikes a crucial balance between performance and efficiency, proving its practicality for real-time operation on UAVs. This makes MATrack a highly valuable solution for real-world low-light surveillance.

\bibliographystyle{IEEEtran}
\balance
\bibliography{icra}

\begin{thebibliography}{10}
\providecommand{\url}[1]{#1}
\csname url@rmstyle\endcsname
\providecommand{\newblock}{\relax}
\providecommand{\bibinfo}[2]{#2}
\providecommand\BIBentrySTDinterwordspacing{\spaceskip=0pt\relax}
\providecommand\BIBentryALTinterwordstretchfactor{4}
\providecommand\BIBentryALTinterwordspacing{\spaceskip=\fontdimen2\font plus
\BIBentryALTinterwordstretchfactor\fontdimen3\font minus \fontdimen4\font\relax}
\providecommand\BIBforeignlanguage[2]{{%
\expandafter\ifx\csname l@#1\endcsname\relax
\typeout{** WARNING: IEEEtran.bst: No hyphenation pattern has been}%
\typeout{** loaded for the language `#1'. Using the pattern for}%
\typeout{** the default language instead.}%
\else
\language=\csname l@#1\endcsname
\fi
#2}}

\bibitem{lei2023multi}
X.~Lei, X.~Hu, G.~Wang, and H.~Luo, ``A multi-uav deployment method for border patrolling based on stackelberg game,'' \emph{Journal of Systems Engineering and Electronics}, vol.~34, no.~1, pp. 99--116, 2023.

\bibitem{al2019appearance}
A.~Al-Kaff, M.~J. G{\'o}mez-Silva, F.~M. Moreno, A.~De~La~Escalera, and J.~M. Armingol, ``An appearance-based tracking algorithm for aerial search and rescue purposes,'' \emph{Sensors}, vol.~19, no.~3, p. 652, 2019.

\bibitem{tian2011video}
B.~Tian, Q.~Yao, Y.~Gu, K.~Wang, and Y.~Li, ``Video processing techniques for traffic flow monitoring: A survey,'' in \emph{2011 14th international IEEE conference on intelligent transportation systems (ITSC)}.\hskip 1em plus 0.5em minus 0.4em\relax IEEE, 2011, pp. 1103--1108.

\bibitem{cao2023towards}
Z.~Cao, Z.~Huang, L.~Pan, S.~Zhang, Z.~Liu, and C.~Fu, ``Towards real-world visual tracking with temporal contexts,'' \emph{IEEE Transactions on Pattern Analysis and Machine Intelligence}, 2023.

\bibitem{he2016deep}
K.~He, X.~Zhang, S.~Ren, and J.~Sun, ``Deep residual learning for image recognition,'' in \emph{Proceedings of the IEEE conference on computer vision and pattern recognition}, 2016, pp. 770--778.

\bibitem{shen2024overlapped}
L.~Shen, X.~Fan, and H.~Li, ``Overlapped trajectory-enhanced visual tracking,'' \emph{IEEE Transactions on Circuits and Systems for Video Technology}, 2024.

\bibitem{vaswani2017attention}
A.~Vaswani, ``Attention is all you need,'' \emph{Advances in Neural Information Processing Systems}, 2017.

\bibitem{cui2022mixformer}
Y.~Cui, C.~Jiang, L.~Wang, and G.~Wu, ``Mixformer: End-to-end tracking with iterative mixed attention,'' in \emph{Proceedings of the IEEE/CVF conference on computer vision and pattern recognition}, 2022, pp. 13\,608--13\,618.

\bibitem{zheng2024odtrack}
Y.~Zheng, B.~Zhong, Q.~Liang, Z.~Mo, S.~Zhang, and X.~Li, ``Odtrack: Online dense temporal token learning for visual tracking,'' in \emph{Proceedings of the AAAI conference on artificial intelligence}, vol.~38, 2024, pp. 7588--7596.

\bibitem{xie2023videotrack}
F.~Xie, L.~Chu, J.~Li, Y.~Lu, and C.~Ma, ``Videotrack: Learning to track objects via video transformer,'' in \emph{Proceedings of the IEEE/CVF conference on computer vision and pattern recognition}, 2023, pp. 22\,826--22\,835.

\bibitem{fu2024prompt}
C.~Fu, Y.~Wang, L.~Yao, G.~Zheng, H.~Zuo, and J.~Pan, ``Prompt-driven temporal domain adaptation for nighttime uav tracking,'' in \emph{2024 IEEE/RSJ International Conference on Intelligent Robots and Systems (IROS)}.\hskip 1em plus 0.5em minus 0.4em\relax IEEE, 2024, pp. 9706--9713.

\bibitem{fu2022highlightnet}
C.~Fu, H.~Dong, J.~Ye, G.~Zheng, S.~Li, and J.~Zhao, ``Highlightnet: highlighting low-light potential features for real-time uav tracking,'' in \emph{2022 IEEE/RSJ International Conference on Intelligent Robots and Systems (IROS)}.\hskip 1em plus 0.5em minus 0.4em\relax IEEE, 2022, pp. 12\,146--12\,153.

\bibitem{ye2021darklighter}
J.~Ye, C.~Fu, G.~Zheng, Z.~Cao, and B.~Li, ``Darklighter: Light up the darkness for uav tracking,'' in \emph{2021 IEEE/RSJ International Conference on Intelligent Robots and Systems (IROS)}.\hskip 1em plus 0.5em minus 0.4em\relax IEEE, 2021, pp. 3079--3085.

\bibitem{ye2022unsupervised}
J.~Ye, C.~Fu, G.~Zheng, D.~P. Paudel, and G.~Chen, ``Unsupervised domain adaptation for nighttime aerial tracking,'' in \emph{Proceedings of the IEEE/CVF conference on computer vision and pattern recognition}, 2022, pp. 8896--8905.

\bibitem{fu2024sam}
C.~Fu, L.~Yao, H.~Zuo, G.~Zheng, and J.~Pan, ``Sam-da: Uav tracks anything at night with sam-powered domain adaptation,'' in \emph{2024 International Conference on Advanced Robotics and Mechatronics (ICARM)}.\hskip 1em plus 0.5em minus 0.4em\relax IEEE, 2024, pp. 31--38.

\bibitem{zhu2024dcpt}
J.~Zhu, H.~Tang, Z.-Q. Cheng, J.-Y. He, B.~Luo, S.~Qiu, S.~Li, and H.~Lu, ``Dcpt: Darkness clue-prompted tracking in nighttime uavs,'' in \emph{2024 IEEE International Conference on Robotics and Automation (ICRA)}.\hskip 1em plus 0.5em minus 0.4em\relax IEEE, 2024, pp. 7381--7388.

\bibitem{wu2024mambanut}
Y.~Wu, X.~Yang, X.~Wang, H.~Ye, D.~Zeng, and S.~Li, ``Mambanut: Nighttime uav tracking via mamba and adaptive curriculum learning,'' \emph{arXiv preprint arXiv:2412.00626}, 2024.

\bibitem{li2025darter}
X.~Li, X.~Li, and S.~Hu, ``Darter: Dynamic adaptive representation tracker for nighttime uav tracking,'' in \emph{Proceedings of the 2025 International Conference on Multimedia Retrieval}, 2025, pp. 1998--2002.

\bibitem{li2022all}
B.~Li, C.~Fu, F.~Ding, J.~Ye, and F.~Lin, ``All-day object tracking for unmanned aerial vehicle,'' \emph{IEEE Transactions on Mobile Computing}, vol.~22, no.~8, pp. 4515--4529, 2022.

\bibitem{wei2023autoregressive}
X.~Wei, Y.~Bai, Y.~Zheng, D.~Shi, and Y.~Gong, ``Autoregressive visual tracking,'' in \emph{Proceedings of the IEEE/CVF Conference on Computer Vision and Pattern Recognition}, 2023, pp. 9697--9706.

\bibitem{bai2024artrackv2}
Y.~Bai, Z.~Zhao, Y.~Gong, and X.~Wei, ``Artrackv2: Prompting autoregressive tracker where to look and how to describe,'' in \emph{Proceedings of the IEEE/CVF conference on computer vision and pattern recognition}, 2024, pp. 19\,048--19\,057.

\bibitem{ye2022joint}
B.~Ye, H.~Chang, B.~Ma, S.~Shan, and X.~Chen, ``Joint feature learning and relation modeling for tracking: A one-stream framework,'' in \emph{European conference on computer vision}.\hskip 1em plus 0.5em minus 0.4em\relax Springer, 2022, pp. 341--357.

\bibitem{kang2025exploring}
B.~Kang, X.~Chen, S.~Lai, Y.~Liu, Y.~Liu, and D.~Wang, ``Exploring enhanced contextual information for video-level object tracking,'' in \emph{Proceedings of the AAAI Conference on Artificial Intelligence}, vol.~39, 2025, pp. 4194--4202.

\bibitem{shi2024explicit}
L.~Shi, B.~Zhong, Q.~Liang, N.~Li, S.~Zhang, and X.~Li, ``Explicit visual prompts for visual object tracking,'' in \emph{Proceedings of the AAAI Conference on Artificial Intelligence}, vol.~38, 2024, pp. 4838--4846.

\bibitem{dong2024loretrack}
S.~Dong, Y.~Feng, Q.~Yang, Y.~Lin, and H.~Fan, ``Loretrack: efficient and accurate low-resolution transformer tracking,'' \emph{arXiv preprint arXiv:2405.17660}, 2024.

\bibitem{ye2022tracker}
J.~Ye, C.~Fu, Z.~Cao, S.~An, G.~Zheng, and B.~Li, ``Tracker meets night: A transformer enhancer for uav tracking,'' \emph{IEEE Robotics and Automation Letters}, vol.~7, no.~2, pp. 3866--3873, 2022.

\bibitem{li2021adtrack}
B.~Li, C.~Fu, F.~Ding, J.~Ye, and F.~Lin, ``Adtrack: Target-aware dual filter learning for real-time anti-dark uav tracking,'' in \emph{2021 IEEE international conference on robotics and automation (ICRA)}.\hskip 1em plus 0.5em minus 0.4em\relax IEEE, 2021, pp. 496--502.

\bibitem{liu2024vmamba}
Y.~Liu, Y.~Tian, Y.~Zhao, H.~Yu, L.~Xie, Y.~Wang, Q.~Ye, and Y.~Liu, ``Vmamba: Visual state space model,'' \emph{arXiv preprint arXiv:2401.10166}, 2024.

\bibitem{xu2022groupvit}
J.~Xu, S.~De~Mello, S.~Liu, W.~Byeon, T.~Breuel, J.~Kautz, and X.~Wang, ``Groupvit: Semantic segmentation emerges from text supervision,'' in \emph{Proceedings of the IEEE/CVF conference on computer vision and pattern recognition}, 2022, pp. 18\,134--18\,144.

\bibitem{gevorgyan2022siou}
Z.~Gevorgyan, ``Siou loss: More powerful learning for bounding box regression,'' \emph{arXiv preprint arXiv:2205.12740}, 2022.

\bibitem{guo2020siamcar}
D.~Guo, J.~Wang, Y.~Cui, Z.~Wang, and S.~Chen, ``Siamcar: Siamese fully convolutional classification and regression for visual tracking,'' in \emph{Proceedings of the IEEE/CVF conference on computer vision and pattern recognition}, 2020, pp. 6269--6277.

\bibitem{zhang2020ocean}
Z.~Zhang, H.~Peng, J.~Fu, B.~Li, and W.~Hu, ``Ocean: Object-aware anchor-free tracking,'' in \emph{European Conference on Computer Vision (ECCV)}, 2020.

\bibitem{cao2021hift}
Z.~Cao, C.~Fu, J.~Ye, B.~Li, and Y.~Li, ``Hift: Hierarchical feature transformer for aerial tracking,'' in \emph{2021 IEEE/CVF International Conference on Computer Vision (ICCV)}, 2021, pp. 15\,457--15\,466.

\bibitem{cao2021siamapn++}
------, ``Siamapn++: Siamese attentional aggregation network for real-time uav tracking,'' in \emph{2021 IEEE/RSJ international conference on intelligent robots and systems (IROS)}.\hskip 1em plus 0.5em minus 0.4em\relax IEEE, 2021, pp. 3086--3092.

\bibitem{Cao2022TCTrackTC}
Z.~Cao, Z.~Huang, L.~Pan, S.~Zhang, Z.~Liu, and C.~Fu, ``Tctrack: Temporal contexts for aerial tracking,'' \emph{2022 IEEE/CVF Conference on Computer Vision and Pattern Recognition (CVPR)}, pp. 14\,778--14\,788, 2022.

\bibitem{zhao2023representation}
H.~Zhao, D.~Wang, and H.~Lu, ``Representation learning for visual object tracking by masked appearance transfer,'' in \emph{2023 IEEE/CVF Conference on Computer Vision and Pattern Recognition (CVPR)}, 2023, pp. 18\,696--18\,705.

\bibitem{Kang2023ExploringLH}
\BIBentryALTinterwordspacing
B.~Kang, X.~Chen, D.~Wang, H.~Peng, and H.~Lu, ``Exploring lightweight hierarchical vision transformers for efficient visual tracking,'' \emph{2023 IEEE/CVF International Conference on Computer Vision (ICCV)}, pp. 9578--9587, 2023. [Online]. Available: \url{https://api.semanticscholar.org/CorpusID:260887522}
\BIBentrySTDinterwordspacing

\bibitem{li2023adaptive}
S.~Li, Y.~Yang, D.~Zeng, and X.~Wang, ``Adaptive and background-aware vision transformer for real-time uav tracking,'' in \emph{Proceedings of the IEEE/CVF International Conference on Computer Vision}, 2023, pp. 13\,989--14\,000.

\bibitem{yao2023sgdvit}
L.~Yao, C.~Fu, and et~al, ``Sgdvit: Saliency-guided dynamic vision transformer for uav tracking,'' \emph{arXiv preprint arXiv:2303.04378}, 2023.

\bibitem{lilearningicml}
Y.~Li, M.~Liu, Y.~Wu, X.~Wang, X.~Yang, and S.~Li, ``Learning adaptive and view-invariant vision transformer for real-time uav tracking,'' in \emph{Forty-first International Conference on Machine Learning}, 2024.

\bibitem{fan2019lasot}
H.~Fan, L.~Lin, F.~Yang, P.~Chu, G.~Deng, S.~Yu, H.~Bai, Y.~Xu, C.~Liao, and H.~Ling, ``Lasot: A high-quality benchmark for large-scale single object tracking,'' in \emph{Proceedings of the IEEE/CVF conference on computer vision and pattern recognition}, 2019, pp. 5374--5383.

\bibitem{huang2019got}
L.~Huang, X.~Zhao, and K.~Huang, ``Got-10k: A large high-diversity benchmark for generic object tracking in the wild,'' \emph{IEEE transactions on pattern analysis and machine intelligence}, vol.~43, no.~5, pp. 1562--1577, 2019.

\bibitem{lin2014microsoft}
T.-Y. Lin, M.~Maire, S.~Belongie, J.~Hays, P.~Perona, D.~Ramanan, P.~Doll{\'a}r, and C.~L. Zitnick, ``Microsoft coco: Common objects in context,'' in \emph{European Conference on Computer Vision (ECCV)}, 2014.

\bibitem{muller2018trackingnet}
M.~Muller, A.~Bibi, S.~Giancola, S.~Alsubaihi, and B.~Ghanem, ``Trackingnet: A large-scale dataset and benchmark for object tracking in the wild,'' in \emph{Proceedings of the European conference on computer vision (ECCV)}, 2018, pp. 300--317.

\bibitem{yu2020bdd100k}
F.~Yu, H.~Chen, X.~Wang, W.~Xian, Y.~Chen, F.~Liu, V.~Madhavan, and T.~Darrell, ``Bdd100k: A diverse driving dataset for heterogeneous multitask learning,'' in \emph{Proceedings of the IEEE/CVF conference on computer vision and pattern recognition}, 2020, pp. 2636--2645.

\bibitem{sun2022shift}
T.~Sun, M.~Segu, J.~Postels, Y.~Wang, L.~Van~Gool, B.~Schiele, F.~Tombari, and F.~Yu, ``Shift: a synthetic driving dataset for continuous multi-task domain adaptation,'' in \emph{Proceedings of the IEEE/CVF Conference on Computer Vision and Pattern Recognition}, 2022, pp. 21\,371--21\,382.

\bibitem{loh2019getting}
Y.~P. Loh and C.~S. Chan, ``Getting to know low-light images with the exclusively dark dataset,'' \emph{Computer Vision and Image Understanding}, vol. 178, pp. 30--42, 2019.

\bibitem{loshchilov2017decoupled}
I.~Loshchilov, ``Decoupled weight decay regularization,'' \emph{arXiv preprint arXiv:1711.05101}, 2017.

\bibitem{li2019siamrpn++}
B.~Li, W.~Wu, Q.~Wang, F.~Zhang, J.~Xing, and J.~Yan, ``Siamrpn++: Evolution of siamese visual tracking with very deep networks,'' in \emph{Proceedings of the IEEE/CVF conference on computer vision and pattern recognition}, 2019, pp. 4282--4291.

\bibitem{fu2021siamese}
C.~Fu, Z.~Cao, Y.~Li, J.~Ye, and C.~Feng, ``Siamese anchor proposal network for high-speed aerial tracking,'' in \emph{2021 IEEE International Conference on Robotics and Automation (ICRA)}.\hskip 1em plus 0.5em minus 0.4em\relax IEEE, 2021, pp. 510--516.

\bibitem{li2018high}
B.~Li, J.~Yan, W.~Wu, Z.~Zhu, and X.~Hu, ``High performance visual tracking with siamese region proposal network,'' in \emph{Proceedings of the IEEE conference on computer vision and pattern recognition}, 2018, pp. 8971--8980.

\bibitem{bhat2019learning}
G.~Bhat, M.~Danelljan, L.~V. Gool, and R.~Timofte, ``Learning discriminative model prediction for tracking,'' in \emph{Proceedings of the IEEE/CVF international conference on computer vision}, 2019, pp. 6182--6191.

\bibitem{danelljan2020probabilistic}
M.~Danelljan, ``Probabilistic regression for visual tracking,'' in \emph{Proceedings of the IEEE/CVF conference on computer vision and pattern recognition}, 2020, pp. 7183--7192.

\end{thebibliography}

\end{document}